\documentclass{article} 
\usepackage{nips13submit_e,times}
\usepackage{hyperref}
\usepackage{url}
\usepackage{amsbsy}
\usepackage{amsmath}
\usepackage{amssymb}
\usepackage{subfigure}
\usepackage{graphicx}
\usepackage{todonotes}
\usepackage{algorithm}	
\usepackage{algorithmic}
\usepackage{amsfonts}
\usepackage[square,sort,comma,numbers]{natbib}
\title{Flexible sampling of discrete data correlations without the marginal distributions}

\author{
Alfredo Kalaitzis \\
Department of Statistical Science and CSML\\
University College London\\
\texttt{a.kalaitzis@ucl.ac.uk} \\
\And
Ricardo Silva \\
Department of Statistical Science and CSML\\
University College London\\
\texttt{ricardo@stats.ucl.ac.uk}
}

\newcommand{ \mb 	}[1]{	\mathbf{#1} 		}

\newcommand{ \bmu 	}{ 	\boldsymbol{\mu} 		}

\newcommand{ \bLambda 	}{ 	\mb{\Lambda} 		}

\newcommand{ \Realspace}{ 	\mathbb{R} 			}
\newcommand{ \Normal 	}[2]{ 	\mathcal{N} \left( #1, #2 \right)	}

\newcommand{ \sm 	}{ 	\setminus 				}

\nipsfinalcopy 

\begin{document}

\maketitle

\begin{abstract}
Learning the joint dependence of discrete
variables is a fundamental problem in machine learning, with many
applications including prediction, clustering and dimensionality
reduction. More recently, the framework of copula modeling has gained
popularity due to its modular parameterization of joint distributions.
Among other properties, copulas provide a recipe for combining
flexible models for univariate marginal distributions with parametric
families suitable for potentially high dimensional dependence
structures. More radically, the extended rank likelihood approach of
Hoff (2007) bypasses learning marginal models completely when such
information is ancillary to the learning task at hand as in, e.g.,
standard dimensionality reduction problems or copula parameter
estimation. The main idea is to represent data by their observable
rank statistics, ignoring any other information from the
marginals. Inference is typically done in a Bayesian framework with
Gaussian copulas, and it is complicated by the fact this implies
sampling within a space where the number of constraints increases
quadratically with the number of data points. The result is slow
mixing when using off-the-shelf Gibbs sampling. We present an
efficient algorithm based on recent advances on constrained
Hamiltonian Markov chain Monte Carlo that is simple to implement and
does not require paying for a quadratic cost in sample size.
\end{abstract}

\section{Contribution}

There are many ways of constructing multivariate discrete
distributions: from full contingency tables in the small dimensional
case \citep{bishop:75}, to structured models given by sparsity
constraints \citep{lauritzen:96} and (hierarchies of) latent variable
models \citep{hinton:06}. More recently, the idea of {\it copula modeling}
\citep{nelsen:07} has been combined with such standard building blocks.
Our contribution is a novel algorithm for efficient Markov chain Monte
Carlo (MCMC) for the copula framework introduced by \cite{hoff:07},
extending algorithmic ideas introduced by \cite{pakman:12}.

A copula is a continuous cumulative distribution function (CDF) with
uniformly distributed univariate marginals in the unit interval $[0,
1]$. It complements graphical models and other formalisms that provide
a modular parameterization of joint distributions. The core idea is
simple and given by the following observation: suppose we are given a
(say) bivariate CDF $F(y_1, y_2)$ with marginals $F_1(y_1)$ and
$F_2(y_2)$. This CDF can then be rewritten as $F(F^{-1}_1(F_1(y_1)),
F^{-1}_2(F_2(y_2)))$. The function $C(\cdot, \cdot)$ given by
$F(F^{-1}_1(\cdot), F^{-1}_2(\cdot))$ is a copula. For discrete
distributions, this decomposition is not unique but still well-defined
\citep{nelsen:07}.
Copulas have found numerous applications in statistics and machine
learning since they provide a way of constructing flexible
multivariate distributions by mix-and-matching different copulas with
different univariate marginals. For instance, one can combine flexible
univariate marginals $F_i(\cdot)$ with useful but more constrained
high-dimensional copulas. We will not further motivate the use of
copula models, which has been discussed at length in recent machine
learning publications and conference workshops, and for which
comprehensive textbooks exist \citep[e.g.,][]{joe:97}. For a recent discussion on 
the applications of copulas from a machine learning perspective,
\cite{elidan:13} provides an overview. \cite{kirshner:07} is an early
reference in machine learning. The core idea dates back at
least to the 1950s \citep{nelsen:07}.

In the discrete case, copulas can be difficult to apply: transforming
a copula CDF into a probability mass function (PMF) is computationally
intractable in general. For the continuous case, a common trick goes as follows:
transform variables by defining $a_i \equiv \hat{F}_i(y_i)$ for an
estimate of $F_i(\cdot)$ and then fit a copula density $c(\cdot,
\dots, \cdot)$ to the resulting $a_i$ \citep[e.g.][]{joe:97}.
It is not hard to check this breaks down in the discrete case
\citep{hoff:07}. An alternative is to represent the CDF to PMF
transformation for each data point by a continuous integral on a
bounded space. Sampling methods can then be used. This trick has allowed many applications of the Gaussian
copula to discrete domains. Readers familiar with probit models will
recognize the similarities to models where an underlying latent
Gaussian field is discretized into observable integers as in Gaussian
process classifiers and ordinal regression \citep{rassmilliams:06}.
Such models can be indirectly interpreted as special cases of the Gaussian copula.

In what follows, we describe in Section \ref{sec:hoff} the
Gaussian copula and the general framework for constructing Bayesian
estimators of Gaussian copulas by \cite{hoff:07}, the extended rank
likelihood framework. This framework entails computational issues
which are discussed. A recent general approach for MCMC in constrained
Gaussian fields by \cite{pakman:12} can in principle be directly
applied to this problem as a blackbox, but at a cost that scales quadratically in
sample size and as such it is not practical in general.
Our key contribution is given in Section \ref{sec:HMC_GCERL}. 
An application experiment on the Bayesian Gaussian copula factor model is
performed in Section \ref{sec:application}.
Conclusions are discussed in the final section.

\section{Gaussian copulas and the \textit{extended rank likelihood}}
\label{sec:hoff}

It is not hard to
see that any multivariate Gaussian copula is fully defined by a
correlation matrix $\mathbf C$, since marginal distributions have no
free parameters. In practice, the following equivalent generative model is
used to define a sample $\mathbf U$ according to a Gaussian copula $\mathcal{GC}(\mathbf C)$:

\begin{enumerate}
\item Sample $\mathbf Z$ from a zero mean Gaussian with covariance matrix $\mathbf C$
\item For each $Z_j$, set $U_j = \Phi(z_j)$, where $\Phi(\cdot)$ is the CDF of the standard Gaussian
\end{enumerate}

It is clear that each $U_j$ follows a uniform distribution in $[0,
1]$. To obtain a model for variables $\{y_1, y_2, \dots, y_p\}$ with
marginal distributions $F_j(\cdot)$ and copula $\mathcal{GC}(\mathbf
C)$, one can add the deterministic step $y_j = F^{-1}_j(u_j)$. Now, given
$n$ samples of observed data $\mathbf Y \equiv \{y_1^{(1)}, \dots, y_p^{(1)}, y_1^{(2)}, \dots, y_p^{(n)}\}$,
one is interested on inferring $\mathbf C$ via a Bayesian approach and the posterior distribution
\[
p(\mathbf C, \theta_F\ |\ \mathbf Y) \propto p_{\mathcal{GC}}(\mathbf Y\ |\ \mathbf C, \theta_F)\pi(\mathbf C, \theta_F)
\]
\noindent where $\pi(\cdot)$ is a prior distribution, $\theta_F$ are marginal parameters for each $F_j(\cdot)$, which in general might need to be marginalized
since they will be unknown, and $p_{\mathcal{GC}}(\cdot)$ is
the PMF of a (here discrete) distribution with a Gaussian copula and marginals given by $\theta_F$.

Let $\mathbf Z$ be the underlying latent Gaussians of the
corresponding copula for dataset $\mathbf Y$. Although $\mathbf Y$ is
a deterministic function of $\mathbf Z$, this mapping is not
invertible due to the discreteness of the distribution: each marginal
$F_j(\cdot)$ has jumps. Instead, the reverse mapping only enforces the constraints where
$y_j^{(i_1)} < y_j^{(i_2)}$ implies $z_j^{(i_1)} < z_j^{(i_2)}$.
Based on this observation, \cite{hoff:07} considers the event
$\mathbf Z \in D(\mb{y})$, where $D(\mb{y})$ is the set of values of $\mathbf Z$ in $\mathbb R^{n \times p}$ obeying those constraints, that is
\[
D(\mb{y}) \equiv \left\{ \mathbf Z \in \mathbb{R}^{n \times p}:
\max \left\{z_j^{(k)} s.t.\ y_j^{(k)} < y_j^{(i)}\right\} < z_j^{(i)} < \min\left\{z_j^{(k)} s.t.\ y_j^{(i)} < y_j^{(k)}\right\} \right\}.
\]

Since $\{\mathbf Y = \mathbf y\} \Rightarrow \mathbf Z(\mathbf y) \in D(\mb{y})$, we have
\begin{equation} \label{eq:hoff}
\begin{array}{rcl}
p_{\mathcal{GC}}(\mathbf Y\ |\ \mathbf C, \theta_F) & = &
p_{\mathcal{GC}}(\mathbf Z \in D(\mb{y}), \mathbf Y \ | \ \mathbf C, \theta_F) \\
& = & p_{\mathcal{N}}(\mathbf Z \in D(\mb{y}) \ |\ \mathbf C)\times p_{\mathcal{GC}}(\mathbf Y |\ \mathbf Z \in D(\mb{y}), \mathbf C, \theta_F),
\end{array}
\end{equation}
\noindent the first factor of the last line being that of a zero-mean a Gaussian density function marginalized over $D(\mb{y})$.

The extended rank likelihood is defined by the first factor of (\ref{eq:hoff}).
With this likelihood, inference for $\mathbf C$ is given simply by marginalizing
\begin{equation}
\label{eq:posterior}
p(\mathbf C, \mathbf Z\ |\ \mathbf Y) \propto I(\mathbf Z \in D(\mb{y}))~ p_{\mathcal{N}}(\mathbf Z |\ \mathbf C) ~ \pi(\mathbf C),
\end{equation}

\noindent the first factor of the right-hand side being the usual binary indicator function.

Strictly speaking, this is not a fully Bayesian method since partial
information on the marginals is ignored. Nevertheless, it is possible
to show that under some mild conditions there is information in the
extended rank likelihood to consistently estimate $\mathbf C$
\citep{murray:13}. It has two important properties: first, in many applications
where marginal distributions are nuisance parameters,
this sidesteps any major
assumptions about the shape of $\{F_i(\cdot)\}$ -- applications
include learning the degree of dependence among variables (e.g., to
understand relationships between social indicators as in
\citep{hoff:07} and \citep{murray:13}) and copula-based dimensionality reduction (a
generalization of correlation-based principal component analysis,
e.g., \citep{han:12}); second, MCMC inference in the extended rank
likelihood is conceptually
simpler than with the joint likelihood,
since dropping marginal models will remove complicated entanglements
between $\mathbf C$ and $\theta_F$. Therefore, even if $\theta_F$ is
necessary (when, for instance, predicting missing values of $\mathbf Y$), an
estimate of $\mathbf C$ can be computed separately and will not
depend on the choice of estimator for $\{F_i(\cdot)\}$. The standard
model with a full correlation matrix $\mathbf C$ can be further
refined to take into account structure implied by sparse inverse
correlation matrices \citep{dobra:11} or low rank decompositions via higher-order
latent variable models \citep{murray:13}, among others.
We explore the latter case in section \ref{sec:application}.

An off-the-shelf algorithm for sampling from (\ref{eq:posterior}) is
full Gibbs sampling: first, given $\mathbf Z$, the (full or structured)
correlation matrix $\mathbf C$ can be sampled by standard methods.
More to the point, sampling $\mathbf Z$ is straightforward if for each
variable $j$ and data point $i$ we sample $Z_{j}^{(i)}$ conditioned on
all other variables. The corresponding distribution is an univariate truncated
Gaussian. This is the approach used originally by Hoff.
However, mixing can be severely compromised by the sampling of $\mathbf Z$,
and that is where novel sampling methods can facilitate inference.


\section{Exact HMC for truncated Gaussian distributions} \label{sec:exactHmc}



Hoff's algorithm modifies the positions of all $Z_j^{(i)}$ associated with 
a particular discrete value of $Y_j$, conditioned on the remaining points. 
As the number of data points increases, the spread of the hard boundaries on $Z_j^{(i)}$,
given by data points of $Z_j$ associated with other levels of $Y_j$, increases. This reduces the space
in which variables $Z_j^{(i)}$ can move at a time.

To improve the mixing, we aim to sample from the \emph{joint} Gaussian distribution of all
latent variables ~$Z_{j}^{(i)}~, i=1\dots n$~, conditioned on other columns of the data,
such that the constraints between them are satisfied
and thus the ordering in the observation level is conserved.
Standard Gibbs approaches for sampling from truncated Gaussians reduce the problem to sampling from
univariate truncated Gaussians.
Even though each step is computationally simple, mixing can be slow when strong
correlations are induced by very tight truncation bounds.

In the following, we briefly describe the methodology recently introduced by \citep{pakman:12}
that deals with the problem of sampling from
~$\log p(\mb{x}) \propto  -\frac{1}{2} \mb{x}^\top \mb{M} \mb{x} + \mb{r}^\top \mb{x}$~,
where ~$\mb{x}, \mb{r} \in \Realspace^{n}$~ and $\mb{M}$~ is positive definite, with linear
constraints of the form ~$\mb{f}_j^\top \mb{x} ~\le~ g_j$~,
where ~$\mb{f}_j \in \Realspace^{n},~ j = 1 \dots m,$~ is the normal vector to some linear
boundary in the sample space.

Later in this section we shall describe how this framework can be applied to the Gaussian
copula extended rank likelihood model. More importantly, the observed rank statistics impose
only linear constraints of the form ~$x_{i_1} \le x_{i_2}$~. We shall describe how this
special structure can be exploited to reduce the runtime complexity of the constrained sampler
from $\mathcal{O}(n^2)$ (in the number of observations) to $\mathcal{O}(n)$ in practice.


\subsection{Hamiltonian Monte Carlo for the Gaussian distribution} \label{subsec:hmcForGaussian}

Hamiltonian Monte Carlo (HMC) \cite{neal:10} is a MCMC method that extends the sampling space with 
auxiliary variables so that (ideally) deterministic moves in the joint space brings the 
sampler to potentially far places in the original variable space. Deterministic moves cannot in
general be done, but this is possible in the Gaussian case.

The form of the Hamiltonian for the general $d$-dimensional Gaussian case with mean
~$\bmu$~ and precision matrix ~$\mb{M}$~ is:
\begin{equation}
  H = \frac{1}{2}~ \mb{x}^\top \mb{M} \mb{x} - \mb{r}^\top \mb{x}
  + \frac{1}{2}~ \mb{s}^\top \mb{M}^{-1}\mb{s} ~,
\end{equation}

where ~$\mb{M}$~ is also known in the present context as the \emph{mass} matrix,
~$\mb{r} = \mb{M} \bmu$~ and ~$\mb{s}$~ is the \emph{velocity}.
Both ~$\mb{x}$~ and ~$\mb{s}$~ are Gaussian distributed so this Hamiltonian can be seen (up to a constant) as
the negative log of the product of two independent Gaussian random variables.
The physical interpretation is that of a sum of \emph{potential} and \emph{kinetic}
energy terms, where the \emph{total} energy of the system is conserved.

In a system where this Hamiltonian function is constant, we can exactly compute its evolution
through the pair of differential equations:
\begin{align}
  \dot{\mb{x}} = \nabla_{s} H = \mb{M}^{-1} \mb{s} ~, \quad \quad
  \dot{\mb{s}} = -\nabla_{x} H = - \mb{M} \mb{x} + \mb{r} ~.
\end{align}
These are solved exactly by $\quad \mb{x}(t) = \bmu + \mb{a} \sin(t) + \mb{b} \cos(t) \quad$,~
where ~$\mb{a}$~ and ~$\mb{b}$~ can be identified at initial conditions ~($t=0$)~:
\begin{align}
  \mb{a} = \dot{\mb{x}}(0) = \mb{M}^{-1}s ~, \quad \quad
  \mb{b} = \mb{x}(0) - \bmu ~.
\end{align}

Therefore, the exact HMC algorithm can be summarised as follows:
\begin{itemize}
  \item Initialise the allowed travel time ~$T$~ and some initial position ~$\mb{x}_0$~.
  \item Repeat for HMC samples ~$k = 1 \dots N$
  \begin{enumerate}
    \item Sample ~$\mb{s}_k \sim \Normal{\mb{0}}{\mb{M}}$
    \item Use ~$\mb{s}_k$~ and ~$\mb{x}_{k}$~ to update ~$\mb{a}$~ and ~$\mb{b}$~
    and store the new position at the end of the trajectory
    ~$\mb{x}_{k+1} = \mb{x}(T)$~ as an HMC sample.
  \end{enumerate}
\end{itemize}
It can be easily shown that the Markov chain of sampled positions has the desired equilibrium
distribution ~$\Normal{\bmu}{\mb{M}^{-1}}$~ \citep{pakman:12}.


\subsection{Sampling with linear constraints} \label{subsec:linearConstraints}

\begin{figure*}[ht]
    \vspace{-1ex}
	\centering
    \subfigure{
    	\includegraphics[width=.25\linewidth]{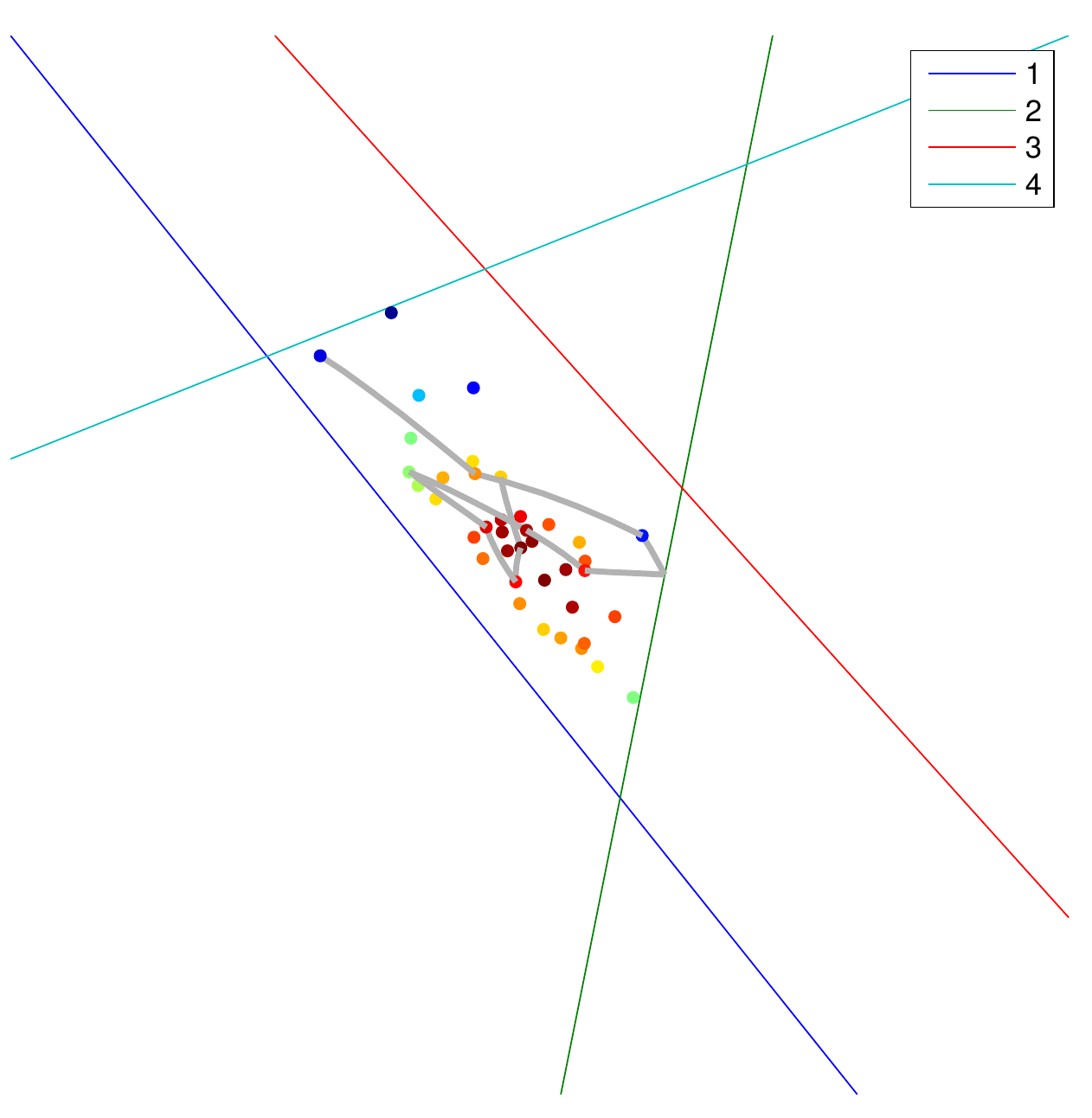}
        \includegraphics[width=.25\linewidth]{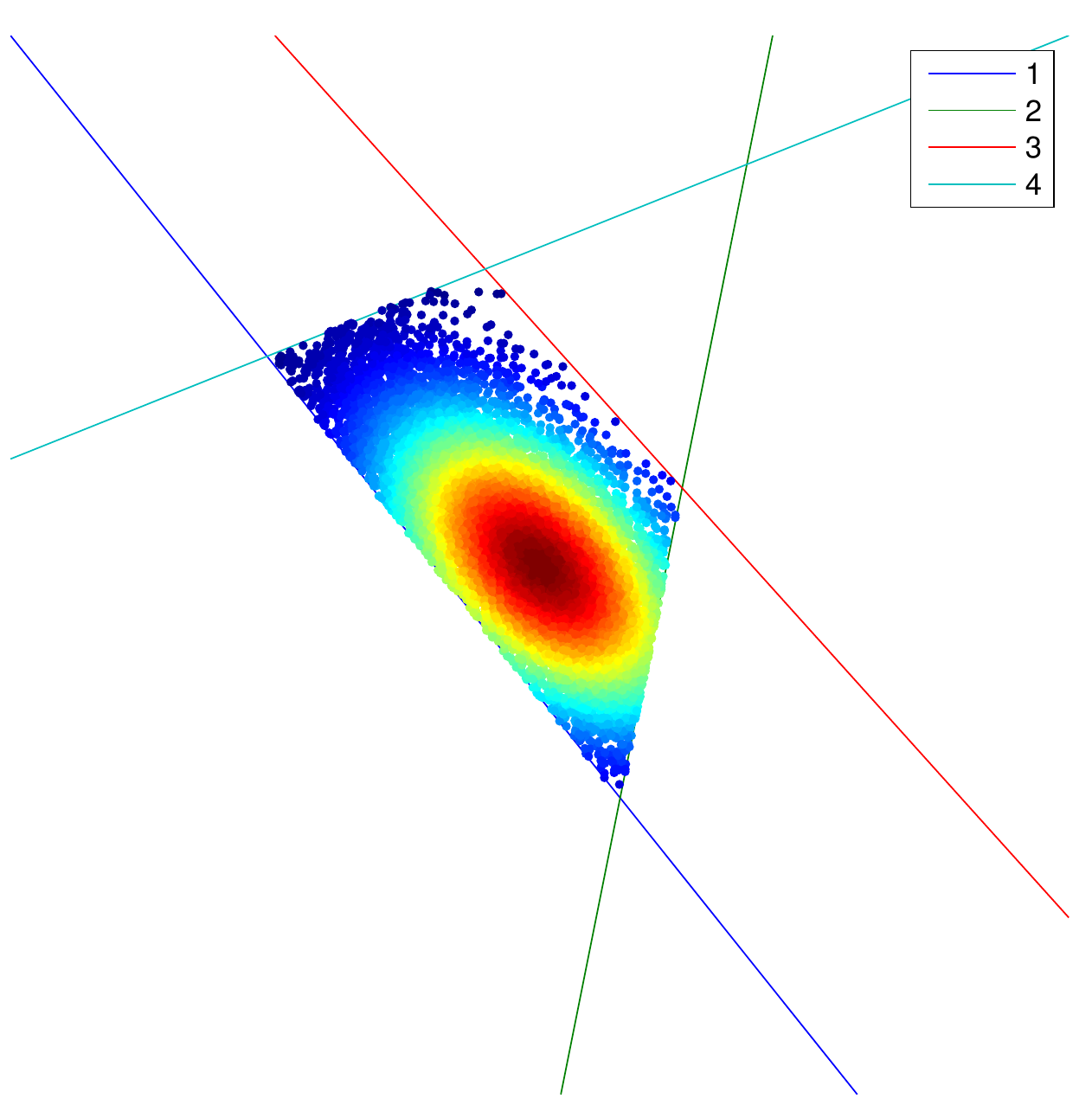}
        \includegraphics[width=.25\linewidth]{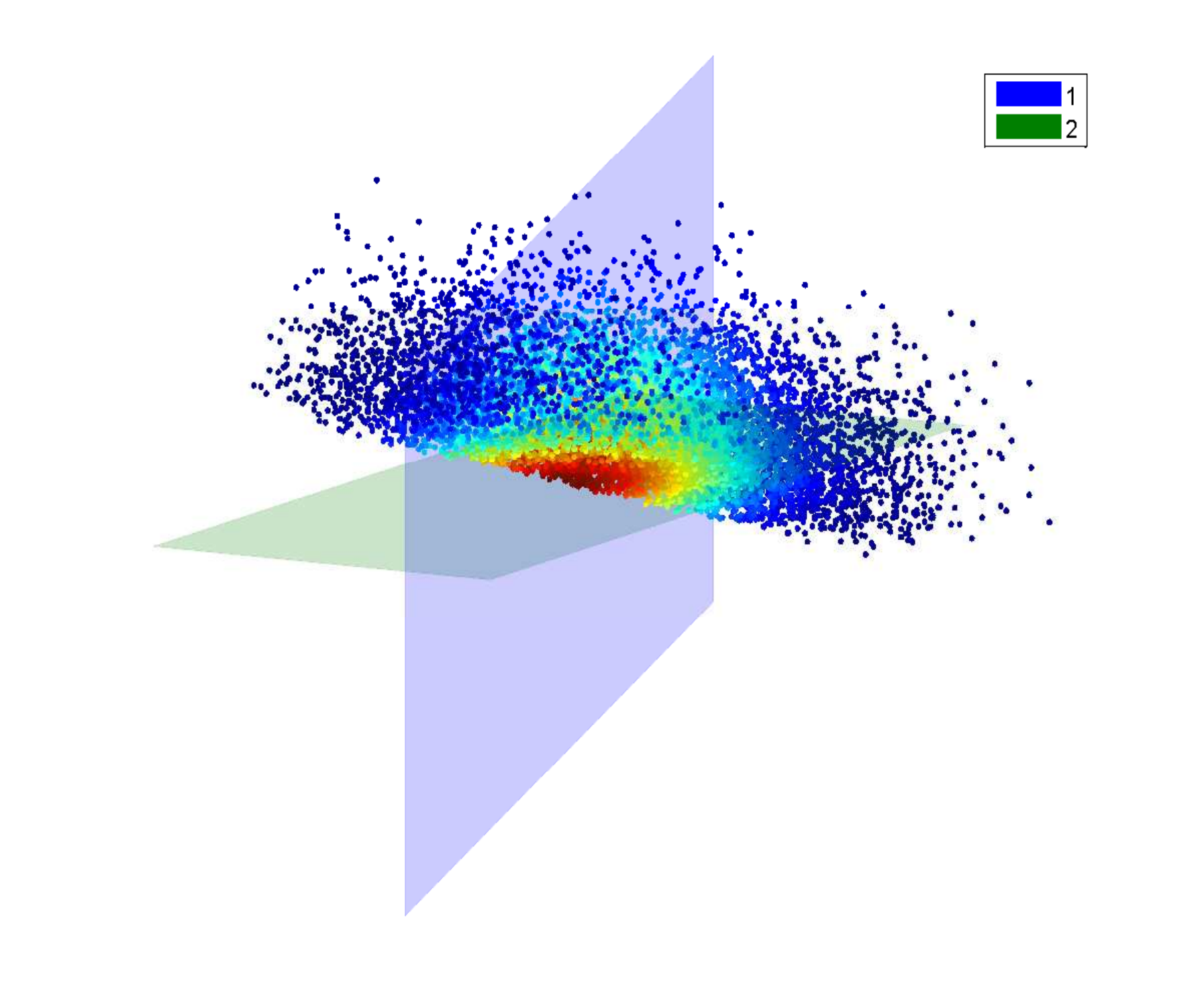}
	}
    \vspace{-1ex}
	\caption{ \label{fig:HMC_examples}
      \textbf{Left}: Trajectories of the first 40 iterations of the exact HMC sampler on a 2D
      truncated Gaussian. A reflection of the velocity can clearly be seen when the particle
      meets wall $\#2$~. Here, the constraint matrix $\mb{F}$ is a ~$4 \times 2$~ matrix.
      \textbf{Center}: The same example after 40000 samples. The coloring of each sample
      indicates its density value. \textbf{Right}: The anatomy of a 3D Gaussian.
      The walls are now planes and in this case $\mb{F}$ is a ~$2 \times 3$~ matrix.
      Figure best seen in color.
	}
    \vspace{-2ex}
\end{figure*}

Sampling from multivariate Gaussians does not require any method as sophisticated
as HMC, but the plot thickens when the target distribution is truncated by linear
constraints of the form
~$\mb{F} \mb{x} \le \mb{g}$~.
Here, ~$\mb{F} \in \Realspace^{m \times n}$~ is a constraint matrix whose every row is the
normal vector to a linear boundary in the sample space.
This is equivalent to sampling from a Gaussian that is confined in the
(not necessarily bounded) convex polyhedron $\{\mb{x} : \mb{F}\mb{x} \le \mb{g}\}$.
In general, to remain within the boundaries of each wall, once a new velocity has been sampled
one must compute all possible collision times with the walls. The smallest of
all collision times signifies the wall that the particle should \emph{bounce} from at
that collision time.
Figure \ref{fig:HMC_examples} illustrates the concept with two simple examples on 2 and 3 dimensions.

The collision times can be computed analytically and their equations can be found in the
supplementary material. We also point the reader to \citep{pakman:12} for a more detailed
discussion of this implementation.
Once the wall to be hit has been found, then position and velocity at impact time are computed
and the velocity is reflected about the boundary normal\footnote{Also equivalent to transforming
the velocity with a \emph{Householder reflection matrix} about the bounding hyperplane.}.
The constrained HMC sampler is summarized follows:
\begin{itemize}
  \item Initialise the allowed travel time ~$T$~ and some initial position ~$\mb{x}_0$~.
  \item Repeat for HMC samples ~$k = 1 \dots N$
  \begin{enumerate}
    \item Sample ~$\mb{s}_k \sim \Normal{\mb{0}}{\mb{M}}$
    \item Use ~$\mb{s}_k$~ and ~$\mb{x}_{k}$~ to update ~$\mb{a}$~ and ~$\mb{b}$~.
    \item Reset remaining travel time ~$T_\mathrm{left} \leftarrow T$~.
    Until no travel time is left or no walls can be reached (no solutions exist), do:
    \begin{enumerate}
      \item Compute impact times with all walls and pick the smallest one, ~$t_h$~
      (if a solution exists).
      \item Compute ~$\mb{v}(t_h)$~ and reflect it about the hyperplane ~$\mb{f}_h$~.
      This is the updated velocity after impact. The updated position is ~$\mb{x}(t_h)$~.
      \item $T_\mathrm{left} \leftarrow T_\mathrm{left} - t_h$~
    \end{enumerate}
    \item Store the new position at the end of the trajectory ~$\mb{x}_{k+1}$~
    as an HMC sample.
  \end{enumerate}
\end{itemize}

In general, all walls are candidates for impact, so the runtime of the sampler is
linear in ~$m$~, the number of constraints. This means that the computational load is
concentrated in step 3(a).
Another consideration is that of the \emph{allocated travel time} ~$T$~.
Depending on the shape of the bounding polyhedron and the number of walls, a very large travel time
can induce many more bounces thus requiring more computations per sample.
On the other hand, a very small travel time explores the distribution more locally
so the mixing of the chain can suffer.
What constitutes a given travel time ``large'' or ``small'' is relative to the dimensionality,
the number of constraints and the structure of the constraints. 

Due to the nature of our problem,
the number of constraints, when explicitly expressed as linear functions,
is $\mathcal{O}(n^2)$~.
Clearly, this restricts any direct application of the HMC framework for Gaussian copula
estimation to small-sample ($n$) datasets.
More importantly, we show how to exploit the structure of
the constraints to reduce the number of candidate walls (prior to each bounce)
to ~$\mathcal{O}(n)$~.


\section{HMC for the Gaussian Copula extended rank likelihood model} \label{sec:HMC_GCERL}

Given some discrete data ~$\mb Y \in \Realspace^{n \times p}$~,
the task is to infer the correlation matrix of the underlying Gaussian copula.
Hoff's sampling algorithm proceeds by alternating between sampling the
\emph{continuous latent} representation ~$Z^{(i)}_j$~ of each ~$Y^{(i)}_j,~$
for $i=1\dots n,~ j=1\dots p$~, and sampling a covariance matrix from an inverse-Wishart
distribution conditioned on the sampled matrix ~$\mb{Z} \in \Realspace^{n \times p}$~,
which is then renormalized as a correlation matrix.

From here on, we use matrix notation for the samples,
as opposed to the random variables -- with $Z_{i, j}$ replacing
$Z_j^{(i)},~ \mb{Z}_{:, j}$ being a column of $\mb{Z}$,~ and $\mb{Z}_{:,\sm j}$
being the submatrix of $\mb{Z}$ without the $j$-th column.

In a similar vein to Hoff's sampling algorithm, we replace the successive sampling of each
~$Z_{i,j}$~ conditioned on
$\mathbf Z_{i,\sm j}$ (a conditional univariate truncated Gaussian) with the simultaneous sampling
of ~$\mb Z_{:,j}$~ conditioned on $\mb Z_{:,\sm j}$.
This is done through an HMC step from a conditional multivariate truncated Gaussian.

The added \emph{benefit} of this HMC step over the standard Gibbs approach,
is that of a \emph{handle} for regulating the \emph{trade-off} between \emph{exploration}
and \emph{runtime} via the allocated travel time $T$. Larger travel times potentially
allow for larger moves in the sample space, but it comes at a cost as explained in the sequel.


\subsection{The \emph{Hough envelope} algorithm} \label{subsec:HoughEnvelope}

\paragraph{The special structure of constraints.}
Recall that the number of constraints is quadratic in the dimension of the distribution.
This is because every ~$\mb{Z}$~ sample must satisfy the conditions of the event
~$\mb{Z} \in D(\mb{y})$~ of the extended rank likelihood (see Section \ref{sec:hoff}).
In other words, for any column ~$Z_{:,j}$~, all entries are organised into a partition ~$L^{(j)}$~
of ~$|L^{(j)}|$~ levels, the number of unique values observed for the discrete or ordinal
variable ~$Y^{(j)}$~.
Thereby, for any two adjacent levels ~$l_k,~ l_{k+1} \in L^{(j)}$~ and any pair
~$i_1 \in l_k,~ i_2 \in l_{k+1}$,~
it must be true that ~$Z_{l_i,j} < Z_{l_{i+1},j}$~.
Equivalently, a constraint ~$\mb{f}$~ exists where ~$f_{i_1} = 1,~ f_{i_2} = -1$~ and ~$g = 0$~.
It is easy to see that ~$\mathcal{O}(n^2)$~ of such constraints are induced by the
order statistics of the ~$j$-th variable.
To deal with this boundary explosion, we developed the \emph{Hough Envelope} algorithm to
search efficiently, within all pairs in ~$\{\mb{Z}_{:,j}\}$,~ in practically \emph{linear} time.

\begin{figure*}[ht]
  \centering
    \begin{minipage}[c]{.4\textwidth}
    \includegraphics[width=1\linewidth]{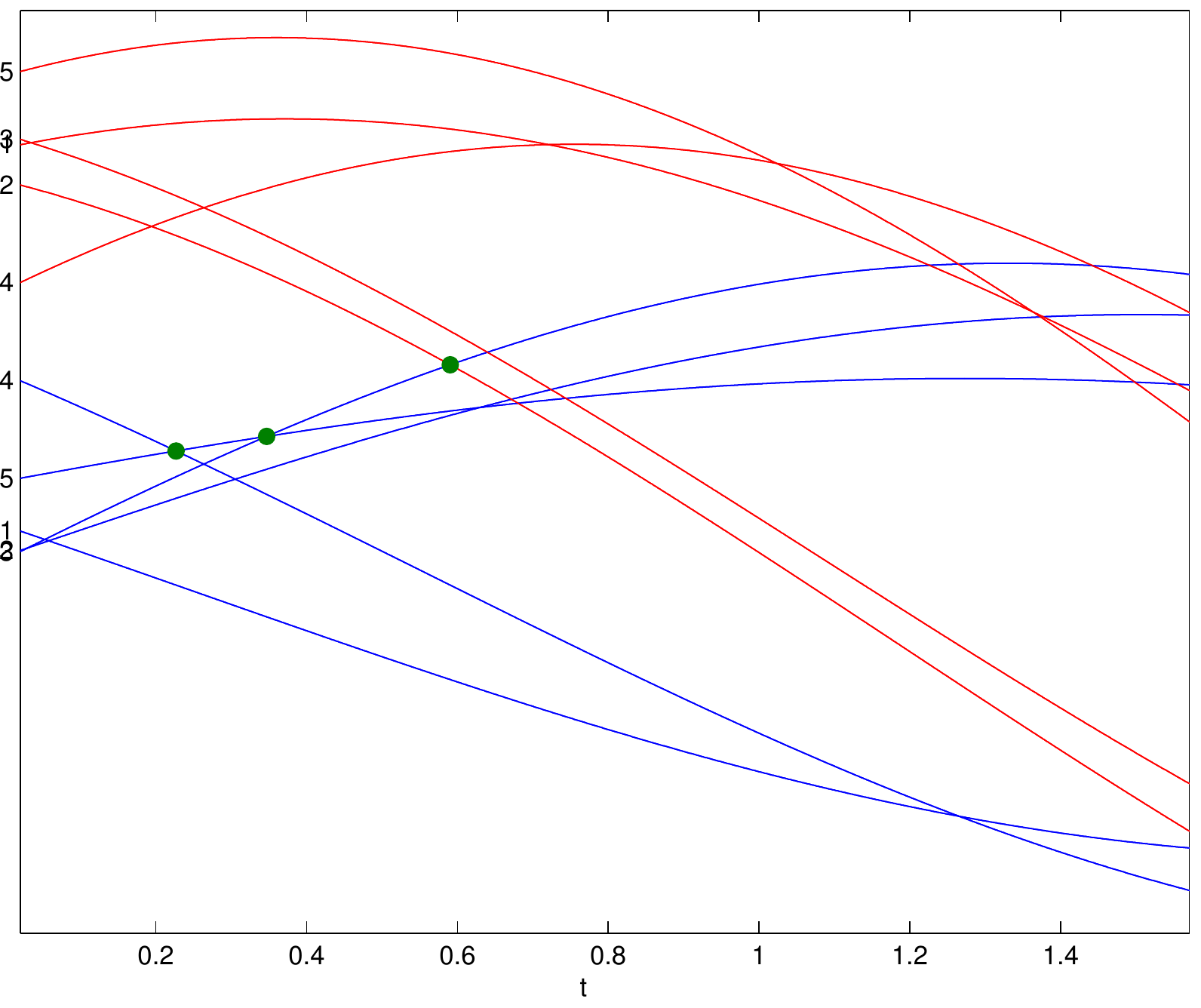}
    \end{minipage}\hfill
    \begin{minipage}[c]{0.5\textwidth}
    \caption{
      The trajectories ~$x_j(t)$~ of each component are sinusoid functions.
      The right-most green dot signifies the wall and the time ~$t_h$~ of the earliest bounce,
      where the first inter-level pair
      (that is, any two components respectively from the blue and red level)
      becomes equal, in this case the constraint \emph{activated} being ~$x_{blue_2} = x_{red_2}$~.
    } \label{fig:HoughEnvelope}
    \end{minipage}
\end{figure*}

Recall in HMC (section \ref{subsec:linearConstraints}) that the trajectory of
the particle, ~$\mb{x}(t)$,~ is decomposed as
\begin{equation}
  x_i(t) = a_i \sin(t) + b_i \cos(t) + \mu_i ~,
\end{equation}
and there are $n$ such functions, grouped into a partition of levels as described above.
The Hough envelope\footnote{The name is inspired from the fact that each ~$x_i(t)$~ is the sinusoid
representation, in \emph{angle-distance} space, of all lines that pass from the ~$(a_i,b_i)$~
point in $a-b$ space. A representation known in image processing as the \emph{Hough transform} \cite{duda:72}.}
is found for every pair of adjacent levels.
We illustrate this with an example of 10 dimensions and two levels in Figure \ref{fig:HoughEnvelope},
without loss of generalization to any number of levels or dimensions.
Assume we represent trajectories for points in level $l_k$ with blue curves, and points in
$l_{k + 1}$ with red curves. Assuming we start with a valid state, at time $t = 0$
all red curves are above all blue curves. The goal is to find the smallest $t$ where a blue
curve meets a red curve. This will be our collision time where a bounce will be necessary.
\begin{enumerate}
  \vspace{-3ex}
  \item First we find the largest component $blue_{max}$ of the blue level at $t=0$.
  This takes ~$\mathcal{O}(n)$~ time.
  Clearly, this will be the largest component until its sinusoid intersects that of
  any other component.
  \item To find the next largest component, compute the roots of ~$x_{blue_{max}}(t) - x_i(t) = 0$~
  for all components and pick the smallest (earliest) one (represented by a green dot).
  This also takes ~$\mathcal{O}(n)$~ time.
  \item Repeat this procedure until a red sinusoid crosses the highest running blue sinusoid.
  When this happens, the time of earliest bounce and its constraint are found.
\end{enumerate}
In the worst-case scenario,
~$n$~ such repetitions have to be made, but in practice we can safely
assume an fixed upper bound $h$ on the number of blue crossings before a inter-level
crossing occurs. In experiments, we found $h << n$, no more than 10 in simulations
with hundreds of thousands of curves. Thus, this search strategy
takes ~$\mathcal{O}(n)$~ time in practice to complete, mirroring the analysis
of other output-sensitive algorithms such as the gift wrapping algorithm for
computing convex hulls \citep{jarvis:73}. Our HMC sampling approach is summarized
in \textbf{Algorithm 1}.

  \begin{algorithm}[!htbp]
    \caption{HMC for GCERL}
    \begin{algorithmic}
      \STATE \# Notation: ~$\mathcal{TMN}(\bmu, \mb{C}, \mb{F})$~
      is a truncated multivariate normal
      with location vector ~$\bmu$,~ scale matrix ~$\mb{C}$~ and constraints encoded by ~$\mb{F}$~
      and ~$\mb{g}=\mb{0}$~.\\
      \# $\mathcal{IW}(df, \mb{V}_0 )$~
      is an \emph{inverse-Wishart} prior with degrees of freedom ~$df$~ and scale matrix ~$\mb{V}_0$~.
      \STATE \textbf{Input:} $\mb{Y} \in \Realspace^{n \times p}$, allocated travel time ~$T$,~
      a starting ~$\mb{Z}$~ and variable covariance ~$\mb{V} \in \Realspace^{p \times p}$~,
      ~$df = p+2,~ \mb{V}_0 = df\mb{I}_p$~ and chain size ~$N$~.
      \STATE Generate constraints ~$\mb{F}^{(j)}$~ from ~$\mb{Y}_{:,j}$~, for ~$j=1 \dots p$~.
      \FOR {samples ~$k = 1 \dots N$}
        \STATE \# Resample ~$\mb{Z}$~ as follows:
        \FOR {variables ~$j = 1 \dots p$}
          \STATE Compute parameters:
          ~$\sigma^2_j =
          \mb{V}_{jj} - \mb{V}_{j,\sm j} \mb{V}_{\sm j, \sm j}^{-1} \mb{V}_{\sm j, j}$~,
          \quad ~$\bmu_j = \mb{Z}_{:,\sm j} \mb{V}_{\sm j,\sm j}^{-1} \mb{V}_{\sm j, j}$~.
          \STATE Get one sample
          ~$\mb{Z}_{:,j} \sim \mathcal{TMN}\left(\bmu_j,~ \sigma^2_j \mb{I},~ \mb{F}^{(j)}\right)$~
          efficiently by using the \emph{Hough Envelope} algorithm, see section
          \ref{subsec:HoughEnvelope}.
        \ENDFOR
        \STATE Resample ~$\mb{V} \sim \mathcal{IW}(df+n, \mb{V}_0 + \mb{Z}^\top \mb{Z})$~.
        \STATE Compute correlation matrix ~$\mb{C}$,~ s.t.
        ~$C_{i,j} = \mb{V}_{i,j} / \sqrt{V_{i,i} V_{j,j}}$~ and store sample, ~$\mb{C}^{(k)} \leftarrow \mb{C}$~.
      \ENDFOR
    \end{algorithmic}
  \end{algorithm}

\vspace{-2ex}
\section{An application on the Bayesian Gausian copula factor model}
\label{sec:application}

\vspace{-1ex}
In this section we describe an experiment that highlights the benefits of our HMC treatment,
compared to a state-of-the-art parameter expansion (PX) sampling scheme.
During this experiment we ask the important question:

\indent \textit{``How do the two schemes compare when we exploit the full-advantage of the HMC
machinery to jointly sample parameters and the augmented data Z, in a model of latent variables
and structured correlations?"}

We argue that under such circumstances the superior convergence speed and mixing of HMC undeniably
compensate for its computational overhead.

\vspace{-1ex}
\paragraph{Experimental setup} \quad
In this section we provide results from an application on the Gaussian copula latent
factor model of \citep{murray:13} (Hoff's model \citep{hoff:07} for low-rank structured correlation matrices).
We modify the parameter expansion (PX) algorithm used by \citep{murray:13} by replacing two
of its Gibbs steps with a single HMC step. We show a much faster convergence to the true mode
with considerable support on its vicinity. We show that unlike the HMC, the PX algorithm falls
short of properly exploring the posterior in any reasonable finite amount of time,
even for small models, even for small samples. Worse, \textit{PX fails in ways one cannot easily detect}.

Namely, we sample each row of the factor loadings matrix $\bLambda$ jointly with the corresponding column of
the augmented data matrix $\mb{Z}$, conditioning on the higher-order latent factors.
This step is analogous to Pakman and Paninski's \citep[][sec.3.1]{pakman:12} use of HMC in the context of a binary
probit model (the extension to many levels in the discrete marginal is straightforward with direct application
of the constraint matrix $\mb{F}$ and the \textit{Hough envelope} algorithm).
The sampling of the higher level latent factors remains identical to \citep{murray:13}.
Our scheme involves no parameter expansion. We do however \textit{interweave} the Gibbs step for the $\mb{Z}$
matrix similarly to Hoff.
This has the added benefit of exploring the $\mb{Z}$ sample space within their current boundaries,
complementing the joint $(\boldsymbol{\lambda}, \mb{z})$ sampling which moves the boundaries jointly.
The value of such "\textit{interweaving}" schemes has been addressed in \cite{Yu:11}.

\vspace{-1.5ex}
\paragraph{Results} \quad
We perform simulations of 10000 iterations, $n=1000$ observations (rows of $\mb{Y}$), travel time $\pi/2$ for HMC
with the setups listed in the following table, along with the elapsed times of each sampling scheme.
These experiments were run on Intel COREi7 desktops with 4 cores and 8GB of RAM.
Both methods were parallelized across the observed variables (p).
\begin{table}[ht] \label{tab:setups}
  \vspace{-2ex}
  \centering
  \begin{tabular}{l|ccc|rrr}
    Figure & p (vars) & k (latent factors) & M (ordinal levels) & elapsed (mins): & HMC & PX \\ 
    \ref{fig:IterVsRMSE_dim20_fa5_obs1000_lvls5} : & 20 & 5 & 2 & &115 &8 \\
    \ref{fig:IterVsRMSE_dim10_fa3_obs1000_lvls2} : & 10 & 3 & 2 & &80 &6 \\
    \ref{fig:IterVsRMSE_dim10_fa3_obs1000_lvls5} : & 10 & 3 & 5 & &203 &16  
  \end{tabular}
  \vspace{-2ex}
\end{table}

\begin{figure}[!htbp]
    \vspace{-2ex}
    \centering
    \subfigure[]{ \label{fig:IterVsRMSE_dim20_fa5_obs1000_lvls5}
      \includegraphics[keepaspectratio=false,height=.06\textwidth, width=.21\textwidth, angle=90, clip=true, trim=80 40 30 50]
      {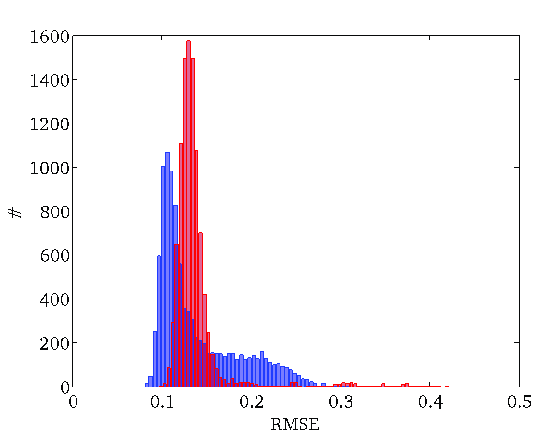}
      \hspace{-1ex}
      \includegraphics[keepaspectratio=false, height=.25\textwidth, width=.27\textwidth]{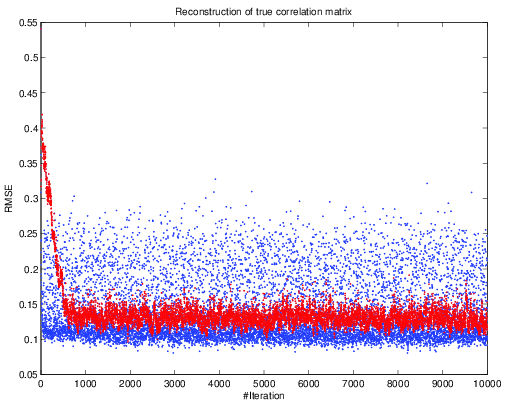}
    }
    \hspace{-3.3ex}
    \subfigure[]{ \label{fig:IterVsRMSE_dim10_fa3_obs1000_lvls2}
	\raisebox{3ex}{
	\includegraphics[keepaspectratio=false,height=.06\textwidth, width=.1\textwidth, angle=90, clip=true, trim=80 40 30 50]
	{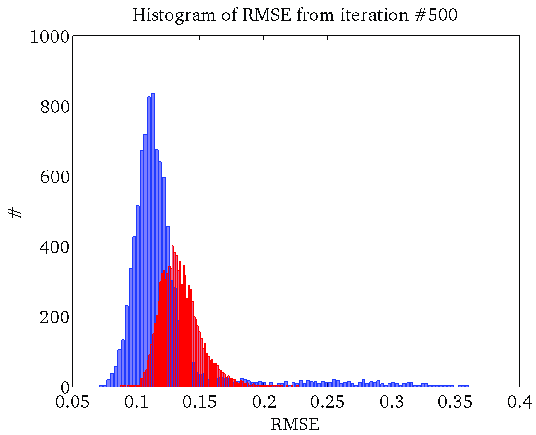}
	}
	\hspace{-1.5ex}
	\includegraphics[keepaspectratio=false, height=.25\textwidth, width=.27\textwidth]{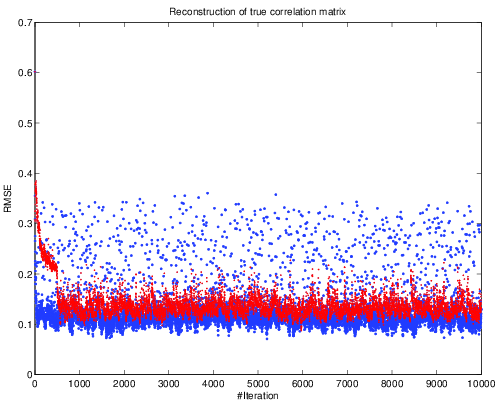}
    }
    \hspace{-3.3ex}
    \subfigure[]{ \label{fig:IterVsRMSE_dim10_fa3_obs1000_lvls5}
	\raisebox{2ex}{
	\includegraphics[keepaspectratio=false,height=.06\textwidth, width=.1\textwidth, angle=90, clip=true, trim=80 40 30 50]
	{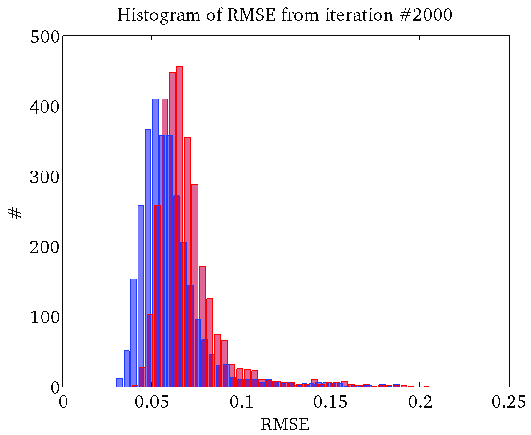}
	}
	\hspace{-1.5ex}
	\includegraphics[keepaspectratio=false, height=.25\textwidth, width=.27\textwidth]{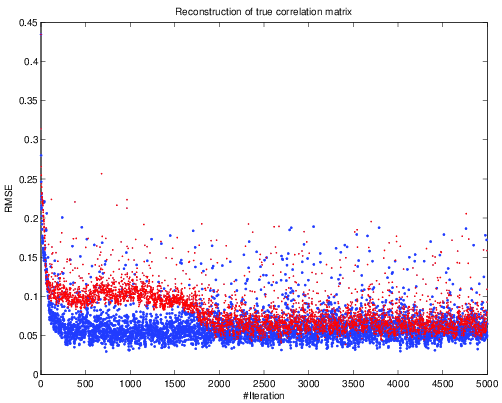}
    }
    \vspace{-2ex}
    \caption{ 
      Reconstruction (RMSE per iteration) of the low-rank structured correlation matrix of the Gaussian copula
      and its histogram (along the left side).\newline
      \textbf{(a)}
      Simulation setup: 20 variables, 5 factors, 5 levels.
      HMC ({\color{blue}blue}) reaches a better mode faster (in \textit{iterations/CPU-time})
      than PX ({\color{red}red}). Even more importantly the RMSE posterior samples of PX
      are concentrated in a much smaller region compared to HMC, even after 10000 iterations.
      This illustrates that PX poorly explores the true distribution.\newline
      \textbf{(b)} Simulation setup: 10 vars, 3 factors, 2 levels.
      We observe behaviors similar to Figure \ref{fig:IterVsRMSE_dim20_fa5_obs1000_lvls5}.
      Note that the histogram counts RMSEs after the burn-in period of PX (iteration \#500).\newline
      \textbf{(c)} Simulation setup: 10 vars, 3 factors, 5 levels.
      We observe behaviors similar to Figures \ref{fig:IterVsRMSE_dim20_fa5_obs1000_lvls5}
      and \ref{fig:IterVsRMSE_dim10_fa3_obs1000_lvls2} but with a thinner tail for HMC.
      Note that the histogram counts RMSEs after the burn-in period of PX (iteration \#2000).
    }
    \vspace{-2ex}
\end{figure}

Many functionals of the loadings matrix $\bLambda$ can be assessed. We focus
on reconstructing the true (low-rank) correlation matrix of the Gaussian copula.
In particular, we summarize the algorithm's outcome with the \textbf{root mean squared error}
\textbf{(RMSE)} of the differences between entries of the ground-truth correlation
matrix and the implied correlation matrix at each iteration of a MCMC scheme (so
the following plots looks like a time-series of 10000 timepoints), see Figures
\ref{fig:IterVsRMSE_dim20_fa5_obs1000_lvls5}, \ref{fig:IterVsRMSE_dim10_fa3_obs1000_lvls2} and
\ref{fig:IterVsRMSE_dim10_fa3_obs1000_lvls5} .

\paragraph{Main message} \quad
HMC reaches a better mode faster (iterations/CPUtime).
Even more importantly the RMSE posterior samples of PX are concentrated in a much smaller region
compared to HMC, even after 10000 iterations. This illustrates that PX poorly explores the true
distribution. As an analogous situation we refer to the top and bottom panels of Figure 14 of
Radford Neal's \textit{slice sampler} paper \citep{neal:03}. If there was no comparison against HMC,
there would be no evidence from the PX plot alone that the algorithm is performing poorly.
This mirrors Radford Neal's statement opening Section 8 of his paper:
\indent \textit{``a wrong answer is obtained without any obvious indication that something is amiss''}.
The concentration on the posterior mode of PX in these simulations is misleading of the truth.
PX might seen a bit simpler to implement, but it seems one cannot avoid using complex algorithms
for complex models. We urge practitioners to revisit their past work with this model to find
out by how much credible intervals of functionals of interest have been overconfident.
Whether trivially or severely, our algorithm offers the first principled approach for checking this out.

\section{Conclusion}
\label{sec:conclusion}

Sampling large random vectors simultaneously in order to improve
mixing is in general a very hard problem, and this is why clever
methods such as HMC or elliptical slice sampling \citep{murray:10b}
are necessary. We expect that the method here developed is useful not
only for those with data analysis problems within the large family of
Gaussian copula extended rank likelihood models, but the method itself
and its behaviour might provide some new insights on MCMC sampling in
constrained spaces in general. Another direction of future work
consists of exploring methods for elliptical copulas, and related
possible extensions of general HMC for non-Gaussian copula models.

\subsection*{Acknowledgements}
The quality of this work has benefited largely from comments by our anonymous
reviewers and useful discussions with Simon Byrne and Vassilios Stathopoulos.
Research was supported by EPSRC grant EP/J013293/1.

\newpage

\bibliography{rbas}

\end{document}